\documentclass[conference]{IEEEtran}
\IEEEoverridecommandlockouts
\usepackage{cite}
\usepackage{amsmath,amssymb,amsfonts}
\usepackage{graphicx}
\usepackage{textcomp}
\usepackage{xcolor}
\usepackage{float}
\usepackage{booktabs}
\usepackage{tabularray,varwidth,enumitem}
\usepackage{multirow, makecell}
\usepackage{algorithm}
\usepackage{algpseudocode}
\usepackage{subcaption}
\usepackage{url}
\usepackage{soul}
\usepackage[switch]{lineno} 
\usepackage{caption} 
\usepackage{hyperref}

\usepackage[normalem]{ulem}
\useunder{\uline}{\ul}{}

\def\BibTeX{{\rm B\kern-.05em{\sc i\kern-.025em b}\kern-.08em
    T\kern-.1667em\lower.7ex\hbox{E}\kern-.125emX}}
\begin{document}

\nolinenumbers


\title{FedMAC: Tackling Partial-Modality Missing in \textbf{Fed}erated Learning with Cross-\textbf{M}odal \textbf{A}ggregation and \textbf{C}ontrastive Regularization}

\author{
\IEEEauthorblockN{
Manh Duong Nguyen\IEEEauthorrefmark{1},
Trung Thanh Nguyen\IEEEauthorrefmark{2},
Huy Hieu Pham\IEEEauthorrefmark{3}, \\
Trong Nghia Hoang\IEEEauthorrefmark{4},
Phi Le Nguyen\IEEEauthorrefmark{1}\IEEEauthorrefmark{6}\thanks{\IEEEauthorrefmark{6}Corresponding author.},
Thanh Trung Huynh\IEEEauthorrefmark{5},
}
\IEEEauthorblockA{
\IEEEauthorrefmark{1} Hanoi University of Science and Technology, Hanoi, Vietnam,
\{duong.nm210243@sis, lenp@soict\}.hust.edu.vn}
\IEEEauthorblockA{\IEEEauthorrefmark{2} Nagoya University, Nagoya, Japan, nguyent@cs.is.i.nagoya-u.ac.jp}
\IEEEauthorblockA{\IEEEauthorrefmark{3} VinUniveristy, Hanoi, Vietnam, hieu.ph@vinuni.edu.vn}
\IEEEauthorblockA{\IEEEauthorrefmark{4} Washington State University, State of Washington, United States, trongnghia.hoang@wsu.edu}
\IEEEauthorblockA{\IEEEauthorrefmark{5} Swiss Federal Institute of Technology Lausanne, Vaud, Switzerland, thanh.huynh@epfl.ch}

}

\maketitle
\begin{abstract}
Federated Learning (FL) is a method for training machine learning models using distributed data sources. 
It ensures privacy by allowing clients to collaboratively learn a shared global model while storing their data locally. 
However, a significant challenge arises when dealing with missing modalities in clients' datasets, where certain features or modalities are unavailable or incomplete, leading to heterogeneous data distribution. While previous studies have tried addressing the issue of complete-modality missing\footnote{{Inter-client missing is when one or more modalities are absent in server and clients' data.}} and intra-client missing\footnote{{Partial missing is when only parts of one or more modalities are absent in server and clients' data.}}, they fail to remains generalization in 
To tackle this challenge, this study proposes a novel framework named FedMAC, designed to address multi-modality missing under conditions of partial-modality missing in FL.
Additionally, to avoid trivial aggregation of multi-modal features, we introduce contrastive-based regularization to impose additional constraints on the latent representation space.
The experimental results demonstrate the effectiveness of FedMAC across various client configurations with statistical heterogeneity, outperforming baseline methods by up to 26\% in severe missing scenarios, highlighting its potential as a solution for the challenge of partially missing modalities in federated systems. Our source code is provided at \href{https://github.com/nmduonggg/PEPSY}{https://github.com/nmduonggg/PEPSY}
\end{abstract}

\begin{IEEEkeywords}
Contrastive Learning, Federated Learning, Modality Missing, Multi-modal Learning.
\end{IEEEkeywords}

\section{Introduction}
\label{sec:introduction}

{
\noindent \textbf{Multi-modal Federated Learning and Modality Missing Issue.} Federated Learning (FL)~\cite{mcmahan2017fedavg} is a decentralized machine learning paradigm that allows multiple devices to train a shared model collaboratively while keeping data localized. 
This approach preserves privacy and reduces data transfer overhead, driven by growing concerns over data privacy and security~\cite{zhang2022data} and the inefficiencies associated with transferring large volumes of data to central servers~\cite{wang2022communication}.
Traditionally, FL methods~\cite{li2019convergence, t2020personalized, li2020federated, fallah2020personalized, cho2022heterogeneous, pham2023sem, nguyen2023cadis, nguyen2022feddrl} have primarily focused on single-modality data (e.g., image or text), which limits their scope of applications. 
With the rapid development of mobile phones and Internet of Things (IoT) devices~\cite{brunete2021smart}, there is increasing potential for leveraging multi-modal data. 
Multi-modal learning integrates various types of data (e.g., sensory, visual, audio), providing a richer and more comprehensive understanding~\cite{kaur2021comparative}. 
Combined with the privacy protection capabilities of FL, multi-modal {FL} enhances the performance and robustness of the system~\cite{xiong2022unified}.
}

\begin{figure}[t]
\centering
\begin{subfigure}{\linewidth}
\centering
    \includegraphics[width=0.95\textwidth]{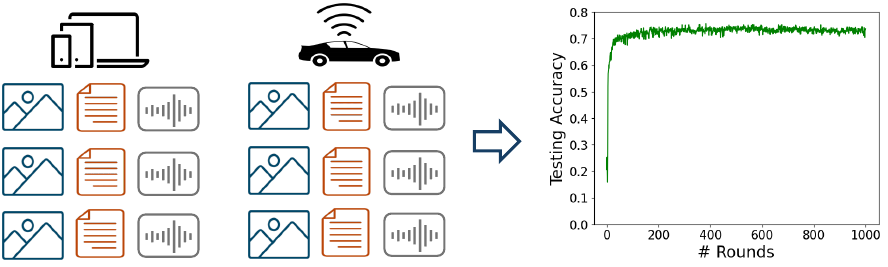}
    \caption{Full-modality and its performance.}
    \label{fig:intro_full_modal}
\end{subfigure}
\hfill
\begin{subfigure}{\linewidth}
\centering
    \includegraphics[width=0.95\textwidth]{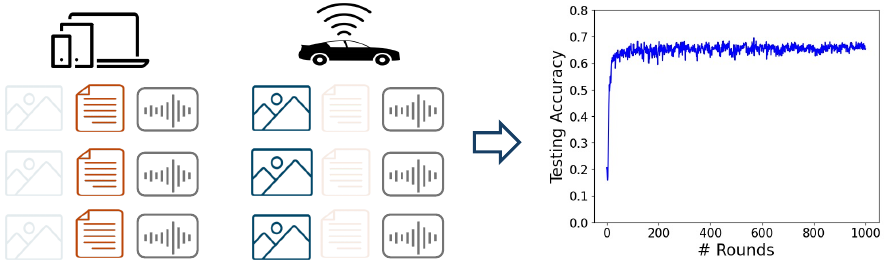}
    \caption{Complete-modality missing and its performance.}
    \label{fig:intro_global_missing}
\end{subfigure}
\hfill
\begin{subfigure}{\linewidth}
\centering
    \includegraphics[width=0.95\textwidth]{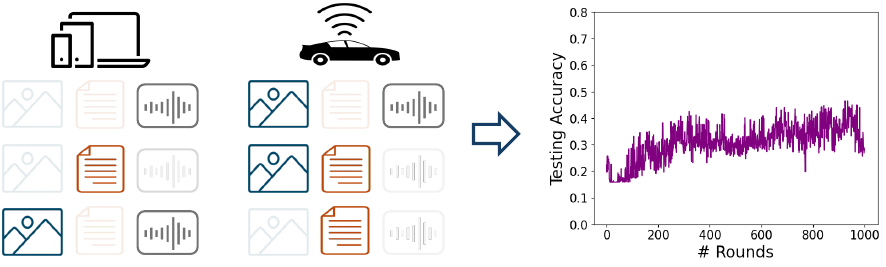}
    \caption{Partial-modality missing and its performance.}
    \label{fig:intro_local_missing}
\end{subfigure}
\caption{Multi-modal settings in Federated Learning and their performance.
In the ideal scenario, full modality performs best. 
However, complete or partial modality missing significantly decreases its performance.}
\label{fig:missing_types}
\vspace{-10pt}
\end{figure}
In the past, {multi-modal FL} methods~\cite{xiong2022unified, chen2024feddat, peng2024fedmm} assume that all modalities are sufficient and complete for every sample. 
This assumption limits the application of FL, as embedded sensors in mobile phones or IoT devices can fail or have poor contact, leading to missing modalities.
{The modality missing in FL can be classified into two types: complete missing and partial missing.
The former refers to a scenario where different combinations of modalities owned by different clients, but the modalities must be complete, while the latter allows diverse modality incompleteness within clients' datasets at an instance level.}
The missing modalities pose significant challenges to multi-modal learning in federated systems. 
As shown in Fig.~\ref{fig:missing_types}, models trained on full modalities experience significant performance degradation when encountering missing modality data {in the testing phase}. \\

\noindent \textbf{Existing Approaches in Dealing with Modality Missing Problems.}
{In conventional centralized learning systems, there are two typical approaches in dealing with modality missing issues: imputation and non-imputation methods.} 
{The core concept of the imputation-based method \mbox{~\cite{zhang2020deep, zhou2022missing}} is to replace the missing modalities with synthetic data, which can be either zeros or the average values of the other modalities.}
However, these methods rely solely on intra-modal information and ignore inter-modal correlations, resulting in suboptimal performance. 
On the other hand, non-imputation methods~\cite{yu2020optimal, chen2020hgmf, poklukar2022geometric} such as hypergraph-based and optimal sparse linear prediction-based approaches, aim to simultaneously benefit models from missing and existing modalities without adding artificial noises into training data.
However, these methods assume centralized training, making them difficult to apply to FL, which relies on decentralized training to ensure privacy.

Despite the critical need for methods to address missing modalities in FL, research in this area remains limited. 
Chen et al.~\cite{chen2022fedmsplit} and Phung et al.~\cite{phung2024mifl} have made early attempts to address this issue in scenarios involving the complete missing of one or multiple modalities, as shown in Fig.~\ref{fig:intro_global_missing}. 
However, as shown in Fig.~\ref{fig:intro_local_missing}, partial-modality missing is more common due to data collection issues or device malfunctions within the network.
Partial-modality missing is more challenging than complete-modality missing, as the server cannot ascertain which devices have missing modalities, resulting in significant noise and fluctuations during training.
Recently, Yu et al.~\cite{yu2024fedinmm} proposed FedInMM to address partial-modality missing. 
However, FedInMM assumes that some time stamps of each missing timeseries-based modality are preserved and employs a Long Short-Term Memory (LSTM)-based module to extract temporal information. 
This assumption limits its generalizability across different scenarios where temporal correlations may be weak or absent.\\

\noindent \textbf{Our Solution.}
{
To overcome the above challenge of training FL under conditions of partial-modality missing, we propose a novel method named FedMAC.
The proposed method introduces modality imputation embeddings that synchronize information between clients and the server, effectively addressing the issue of diverse modalities. 
Additionally, we propose a cross-modal aggregation approach to minimize biases towards specific modalities and capture correlations between different types of inputs. 
This approach reconstructs the representation of a modality by aggregating features from the remaining ones. 
The aggregated features are then combined with the original representations to utilize various sources of information describing the same modality. 
We also implement a contrastive-based regularization to ensure that the aggregation process involves only relevant modalities, preventing trivial feature aggregation.
}

{
The main contributions of this study are as follows:
\begin{itemize}
    \item {We are the first to thoroughly investigate the problem of partially missing modalities in FL.
    To tackle this, we introduce FedMAC, an innovative client-side architecture designed to handle complete and partial modality missing issues.
    FedMAC imputes the missing modalities by leveraging intra-modal information with modality-specific learnable embeddings and inter-modal correlations through a cross-modal reconstruction process.}
    \item We leverage contrastive learning to regularize representations learned from each modality, avoiding trivial cross-modal reconstruction and enhancing inter-modality information. 
    This mechanism enriches the alternative features of the missing ones, thus boosting overall performance.
    \item We conduct extensive experiments across multiple scenarios of missing modalities to demonstrate the effectiveness of the proposed method in real-world conditions. 
    The experimental results show the superiority of the proposed method compared to state-of-the-art methods on the subset of multi-modal PTB-XL dataset~\cite{dataset}. 
\end{itemize}
}

The remainder of this paper is organized as follows. 
Section~\ref{sec:methodology} presents the problem formulation and the proposed FedMAC method for handling partial missing data in multimodal FL.
The experimental settings and the evaluation are discussed in Section~\ref{sec:evaluation}. 
Finally, the conclusion and potential future directions are presented in Section~\ref{sec:conclusion}.


\section{Methodology}
\label{sec:methodology}
\begin{figure}[t]
    \centerline{\includegraphics[width=0.885\linewidth]{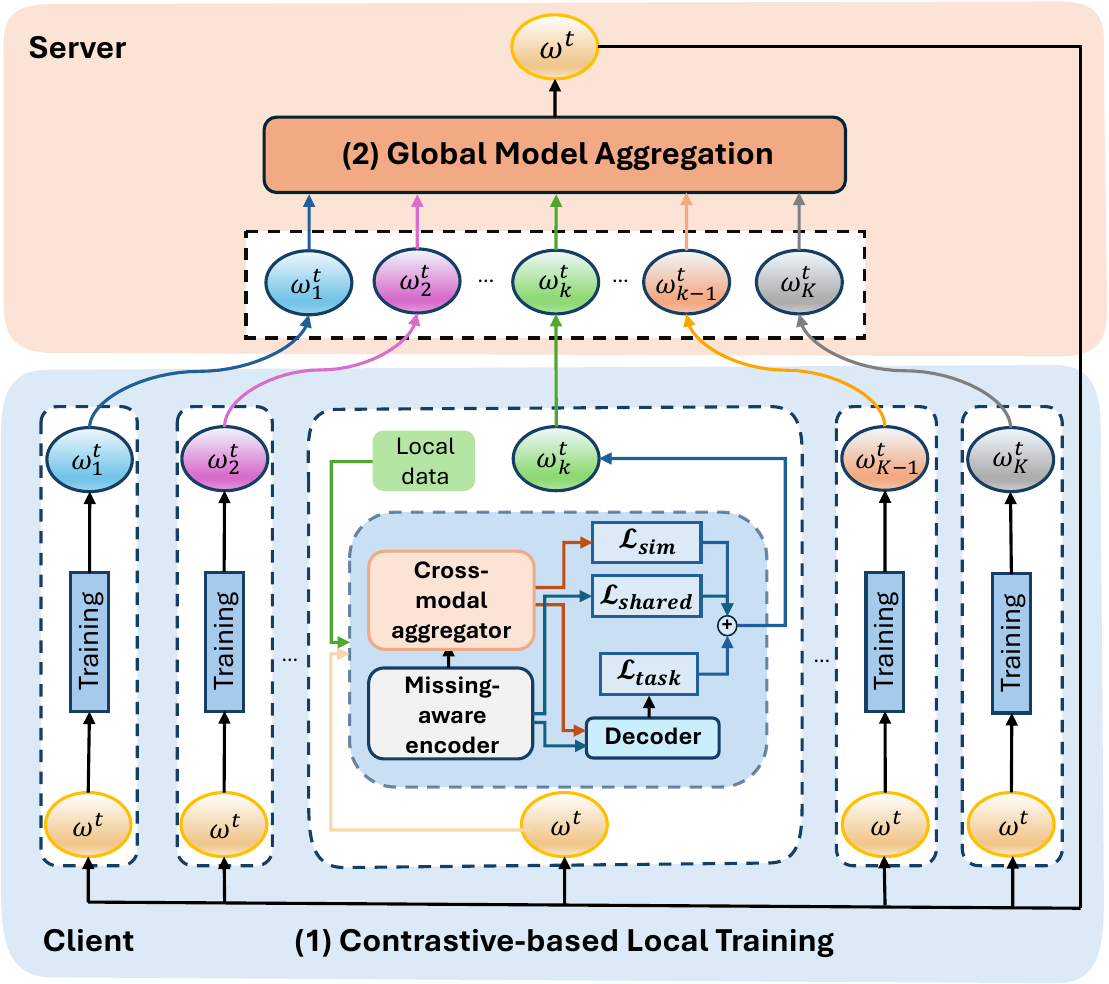}}
    \caption{\textbf{Overview of FedMAC.} We utilize a conventional weighted averaging aggregation on the server side and propose a novel model architecture and training algorithm on the client side.}
    \label{fig:overview_framework}
    \vspace{-10pt}
\end{figure}
\subsection{Problem Formulation}    
\label{subsec:problem_formulation}
In the Federated Learning (FL) system, we assume there are \(K\) clients denoted as $(C_1, \dots, C_K)$.
Each client $C_k$ possesses a local multi-modal dataset $\mathbb{D}_k$, where $\mathbb{D}_k = \{{x}_i\}_{i=1}^{N_k}$ ($N_k$ is the cardinality of $\mathbb{D}_k$). 
Let $\mathcal{M}(x_i)$ represents the modality set of a sample $x_i$, and $\mathcal{M}(\mathbb{D}_k)$ depicts the union of modality sets of all samples belonging to $\mathbb{D}_k$. 
We consider a scenario with occurrence of both \textit{partial-modality missing} (i.e., $\mathcal{M}(x_i)$ varies across samples from the same client), and \textit{complete-modality missing} (i.e., $\mathcal{M}(\mathbb{D}_k)$ differs among clients).
Under such a scenario, the problem asks to train a global model capable of making accurate predictions. The objective is to optimize the following empirical loss:
\begin{equation}
    \underset{w}{\mathrm{argmin}} \frac{1}{\sum_{k=1}^K N_k} \sum_{k=1}^K \sum_{i=1}^{N_k} \mathcal{L} (f(w, {x}_i, y_i)),
\end{equation}
where \(\mathcal{L}\) represents the loss function adopted for the specific problem, \( w \) represents the model parameters, $x_i$ denotes the input data with missing modalities, and \(y_i\) denotes the corresponding labels.

\subsection{Overview of FedMAC}

Federated Learning (FL) consists of two main phases: global model aggregation and client-specific local training. Unlike centralized training methods, FL enables \(K\) clients to train a local model \(\omega_i\) \((i = \{1, \dots, K\})\) on their data for \(E\) epochs and share only the model parameters with a global server. Upon receiving the local models from the clients, the server aggregates them to form a global model. This global model is subsequently distributed back to the clients for the next training round. This process is a single training round and continues until the global model converges.

This study focuses on the local training phase and employs the widely-used FedAvg algorithm~\cite{mcmahan2017fedavg}, which relies on weighted averaging for global model aggregation.
A key challenge in multi-modal FL, where data types or ``modalities" vary across clients, is modality heterogeneity. In simple terms, different clients may have access to distinct sets of modalities and observe incomplete collection of training samples, which can introduce noise and degrade model performance.

To address this challenge, we propose FedMAC, a method focused on extracting modality-invariant features from various input types, thereby eliminating modality-specific biases during training.
Our approach is based on the hypothesis that each modality can be viewed as an augmented version of a core underlying feature. This means that the essential information required for a specific task is embedded within different forms of representation, regardless of their original modality. For instance, an image of a cat and the sound of a ``meow" can be seen as different augmentations of the underlying ``cat feature" that the model should ultimately predict.

The proposed FedMAC includes two main components in the client's local model: the \emph{Missing-Aware Encoder} and the \emph{Cross-Modal Aggregator} module.
The Missing-Aware Encoder extracts shared features across modalities and handles missing data through modality imputation embeddings. 
The Cross-Modal Aggregator reinforces feature invariance by representing each modality as a linear combination of others. 
To prevent the linear aggregation from becoming trivial and ensure meaningful latent representations, we introduce a \emph{Contrastive-based Regularization}. 
Details of the client's local model architecture are provided in Section~\ref{subsec:client_model}, with the contrastive-based training algorithm discussed in Section~\ref{subsec:contrastive_loss}.

\subsection{{Client's Multi-modal Representation Generation}}
\label{subsec:client_model}

\begin{figure}[t]
\centerline{\includegraphics[width=0.9\linewidth]{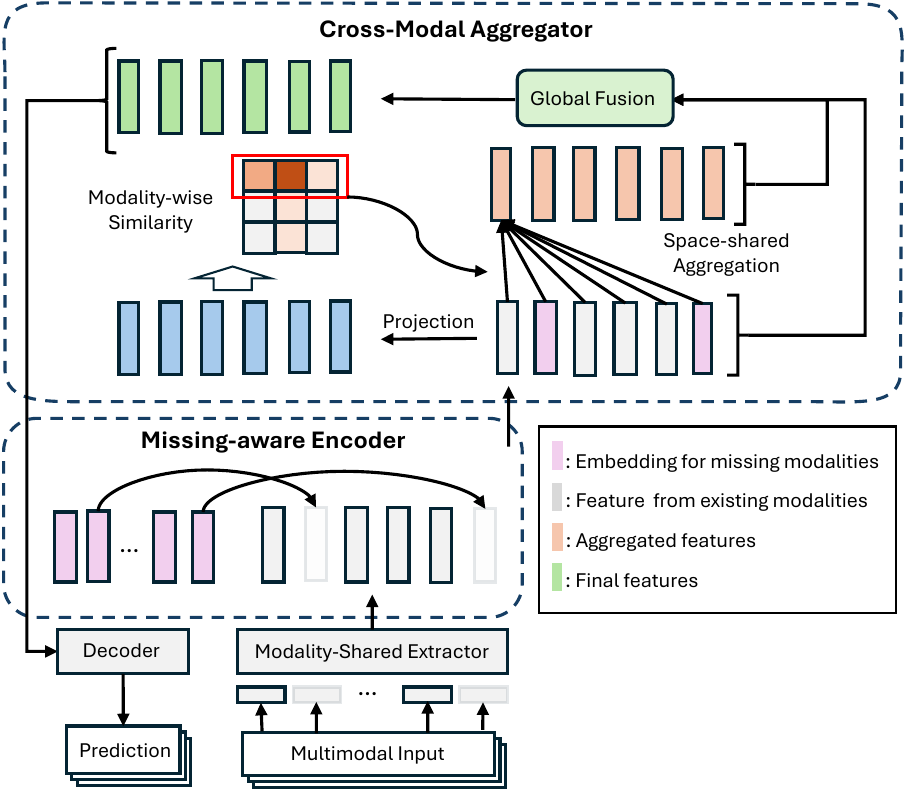}}
    \caption{Details of our proposed architecture for clients' local models.}
    \label{fig:overall_client_model}
    \vspace{-10pt}
\end{figure}
\subsubsection{Missing-Aware Encoder} \label{sec:missing_aware_encoder}
{This module aims to generate modality encoding regardless of input existing modalities. The missing problem is handled by keeping extracted features from modality-shared extractor and utilizing modality imputation embeddings for missing ones.}

\noindent \textbf{Modality-shared Extractor.}
{
For a client $k$, given its local incomplete dataset $\mathbb{D}_k$, the multi-modal sample ${x}_i = \{{x}_i^m\}_{m=1}^{\mathcal{M}_k(x_i)}$ is fed into a modality-shared extractor to extract features for each modality in a modality-shared space $\mathbb{R}^{d_h}$, where $d_h$ is the latent dimension. 
This aims to ensure that the core features relevant to the target task are consistently extracted, regardless of the differences in input modalities.
The features extracted for the sample ${x}_i^m$ are as:
\begin{equation}
    \hat{h}_i^m = \gamma ({x}_i^m),
\end{equation}
where $\hat{h}_i^m \in \mathbb{R}^{d_h}$ represents the output feature of modality $m$, $\gamma$ is the modality-shared function.
}


\noindent \textbf{Modality Imputation Embeddings.}
\label{sec:embeddings}
{
To fully capture intra-modal information and ensure synchronization between clients and the server, we propose a learnable embedding for each modality. 
Specifically, each modality is assigned a learnable embedding $e^m$, which is initialized in the latent space $\mathbb{R}^{d_h}$.
Consequently, a client $k$ possesses an embedding set $\{e^m\}_{m}^{\mathcal{M}({\mathbb{D}_k})}$. 
These embeddings are used to replace missing modalities in the sample ${x}_i$, thereby obtaining more meaningful information. The complete features $h_i^m$ are obtained as:
\begin{equation}
    h_i^m = 
    \begin{cases} 
      \hat{h}_i^m & \text{, if } m \in \mathcal{M}(x_i) \\
      e^m & \text{, otherwise}.
   \end{cases}
\end{equation}
We refer to the set of features $H_i = \{h_i^m\}_{m}^{\mathcal{M}(\mathbb{D}_k)}$ as \textit{space-shared feature}, which is a representation for each sample by missing-aware encoder.
}


\subsubsection{Cross-Modal Aggregator} \label{sec:rebuilder}


Conventional multi-modal learning often directly fuses imputed embeddings with extracted features, limiting model flexibility and inadequately handling diverse modality sets and varying labels. 
To overcome this, we introduce the Cross-Modal Aggregator, which dynamically reformulates missing-modality features based on existing ones and vice versa, allowing for instance-level specialization. 
Our approach maps space-shared features into a higher-level space to learn modality-wise similarities, which guide a weighted aggregation back into the original space.
A global fusion module dynamically fuses features before and after aggregation, enabling adaptive integration across modalities and improving the model's ability to handle diverse multi-modal data.

\noindent \textbf{Modality-wise Similarity.} 
This component leverages the similarities between space-shared representations \( h_i^m \) for weighted aggregation. 
Specifically, we use a simple MLP layer for each modality to generate modality-specific representations and capture high-level similarities between modalities. 
The process is described as follows:
\begin{equation} \label{eq:projector}
z_i^m = \text{MLP}_m(h_i^m),
\end{equation}
where $z_i^m$ represents the modality-wise projection obtained through $\text{MLP}_m$ for modality $m$.

For clarity, let's denote the set of all such modality-wise representations as $Z = \{z_i^m\}_{i, m}$. To formalize the estimation of modality-wise similarities for client $k$, we define the similarity matrix $S_Z$ as follows:
\begin{equation}
S_Z \in \mathbb{R}^{T_k \times T_k}; \quad S^{u, v}_Z = \frac{u \cdot v^T}{||u|| \cdot ||v||}; \quad u, v \in Z,
\end{equation}
where $T_k = \mathcal{M}(\mathbb{D}_k)(N_k + 1)$. Here, $S_Z$ represents the pairwise similarity matrix for modality features of $\mathcal{M}(\mathbb{D}_k)$ of $N_k$ samples in the modality-wise space $Z$, with similarities computed using cosine distance. $S^{u, v}_Z$ captures the similarity between a pair of representations $u$ and $v$ within $Z$.

It is important to note that the similarity matrix $S_Z$ also includes similarities between $\mathcal{M}(\mathbb{D}_k)$ additional embeddings, which are treated as special instances. This inclusion is crucial because if no modality is missing, the information imputed by these extra embeddings might not be learned effectively, potentially introducing noise into the server-side aggregation in federated learning systems. By accounting for these similarities, our approach ensures that the model remains robust and accurate, even in the presence of diverse modality sets and varying ground-truth labels across clients.

\noindent \textbf{Space-shared Aggregation}.
As depicted in Fig.~\ref{fig:overall_client_model}, the cross-modal aggregation process for the $m$-th modality feature of the $i$-th instance within the shared space is driven by the learned modality-wise similarities. The aggregated representation for this feature, denoted as $\tilde{h}_i^m$, is computed as follows:
\begin{equation}    \label{eq:cross-modal-agg}
    \tilde{h}_i^m = \sum_{z', h' \in Z' \times H'} \frac{\exp\left(S_Z^{z_i^m, z'} / \tau\right)}{\sum_{z'' \in Z'} \exp\left(S_Z^{z', z''} / \tau\right)} h',
\end{equation}
where $Z'$ represents all modality-wise representations in the dataset except for the one being reconstructed, and $H'$ includes all modality features except for $h_i^m$. In this context, $z'$ is the modality-wise projection of $h'$, calculated as $z' = \text{MLP}_m(h')$. The resulting aggregated representation, $\tilde{h}_i^m$, is also in the shared space $\mathbb{R}^{d_h}$. The parameter $\tau$ acts as a temperature hyperparameter, controlling the sharpness of the softmax normalization applied to the modality-wise similarities.

Importantly, Eq.~\ref{eq:cross-modal-agg} shows that the reconstruction of each modality feature \(\tilde{h}_i^m\) is based on its corresponding modalities \(\{x_i^m\}_{m}^{\mathcal{M}({\mathbb{D}_k})}\) while also incorporating modality representations from other instances. 
This approach consequently ensures the model to suppress interference from less related modalities and accurately learns the representations for each modality within an instance, thereby improving the overall reconstruction of instance-specialized features.

\begin{figure}[t]
    \centerline{\includegraphics[width=0.95\linewidth]{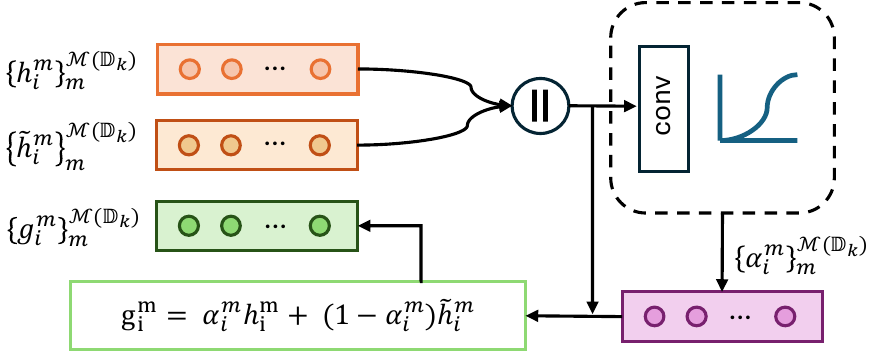}}
    \caption{Global fusion mechanism.}
    \label{fig:global_fusion}
    \vspace{-10pt}
\end{figure}

\noindent \textbf{Global Fusion}. Despite enhancement in similarity learning, excluding $h_i^m$ in the final representation built for $x_i^m$ might decrease expressiveness of feature, especially that of existing modalities with meaningful hidden features. Therefore, as can be seen from Fig.~\ref{fig:global_fusion}, we propose Global Fusion module to fuse the latent features before and after space-shared aggregation with learnable coefficient vector $\alpha \in \mathbb{R}^{d_h}$:
\begin{equation}
    g_i^m = \alpha_i^m h_i^m + (1 - \alpha_i^m) \tilde{h}_i^m,
\end{equation}
in which $\alpha_i^m$ is adopted by a simple convolution network with three convolution layers. By adopting sigmoid activation function, $\alpha_i^m$ represents a linear coefficient corresponding to importance of features before and after aggregation:
\begin{equation}
    \alpha_i^m = \text{Conv}({h}_i^m || \tilde{h}_i^m).
\end{equation}
Finally, a final representation $G_i$ of ${x}_i$ is obtained by concatenating outputs $\{g_i^1, {g}_i^2, \dots, {g}_i^{\mathcal{M}({\mathbb{D}_k})}\}$ and fed into a decoder to generate a prediction.

\subsection{Contrastive-based Regularization}
\label{subsec:contrastive_loss}
In this study, we propose leveraging contrastive loss to enhance the modality-invariant objectives of the missing-aware encoder and guide the learning of modality-wise similarity. 
While missing modalities are addressed through cross-modal aggregation and the assumption of a shared representation space, as outlined in Section~\ref{sec:missing_aware_encoder}, ensuring the model accurately captures similarity within the modality-wise representation space remains challenging without explicit constraints. 
The contrastive regularization aims to calibrate the structure of the modality-wise representation space, hence preventing trivial aggregation and enforcing meaningful learning.

\noindent \textbf{Space-shared Contrastive Loss} ($\mathcal{L}_{shared}$), is designed to ensure that all modality-specific features belonging to a single instance are closely aligned within the shared representation space. This loss functions by defining positive and negative pairs, where positive pairs consist of modality-specific features from the same instance, which should be close in the shared space, and negative pairs comprise features from different instances, which should be pushed apart. Mathematically, this loss is expressed as:
\begin{equation}    
    \label{eq:contrastive_sim_loss}
    \mathcal{L}_{shared} = - \sum_{h \in H} \sum_{h' \in H^+} \log \frac{\exp(R_H^{h, h'} / \tau)}{\sum_{h'' \in H / \{h\}} \exp(R_H^{h, h''} / \tau)},
\end{equation}
where $R_H$ is the similarity matrix in the shared space, $H^+$ is a set of representations desired to be close to $h$, and $\tau$ is a temperature parameter that controls the sharpness of the similarity distribution. This loss ensures that different modalities of the same instance are mapped to similar hidden representations, reinforcing the consistency of cross-modal information.

\noindent \textbf{Modality-Wise Similarity Contrastive Loss} ($\mathcal{L}_{sim}$), targets the preservation of accurate cross-modal relationships within the original modality-specific representation space. This loss also uses positive and negative pairs as similar as $\mathcal{L}_{shared}$, but within the modality-wise space. The loss is defined as:
\begin{equation}    
    \label{eq:contrasitve_shared_loss}
    \mathcal{L}_{sim} = - \sum_{z \in Z} \sum_{z' \in Z^+} \log \frac{\exp(S_Z^{z, z'} / \tau)}{\sum_{z'' \in Z / \{z\}} \exp(S_Z^{z, z''} / \tau)},
\end{equation}
where $S_Z$ is the similarity matrix in the modality-wise space, $Z^+$ is a set of representations desired to be close to $z$. This contrastive loss ensures that the model learns modality-wise representations that accurately reflect cross-modal similarities, maintaining the integrity of each modality's unique features while aligning related features.

Together, these two contrastive losses contribute to the final objective function for training the local models. This function combines the task-specific loss $\mathcal{L}_{task}$ with the contrastive losses, leading to the final objective:
\begin{equation}    
    \label{eq:final_loss}
    \mathcal{L} = \mathcal{L}_{task} + \lambda (\mathcal{L}_{shared} + \mathcal{L}_{sim}),
\end{equation}
where $\lambda$ is a hyperparameter that controls the influence of the contrastive regularization on the final loss. This combined approach ensures that the model effectively learns robust, modality-invariant representations while preserving critical cross-modal relationships.

\subsection{Aggregation Operation on Server}
In this study, FedMAC employs the FedAvg algorithm~\cite{mcmahan2017fedavg} for server aggregation to obtain the global model after each communication round. 
The server initializes the global model \(\omega^0\), and after each round, clients replace their local models with the updated global model. 
Clients then update their models on local data for a few epochs before sending them back to the server. 
The server aggregates the model as follows:
\begin{equation}    
    \label{eq:server_aggregation}
    \omega^t = \sum_{k=1}^K \frac{N_k}{N} \omega_k^t,
\end{equation}
where \(N = \sum_{k=1}^K N_k\) is the total number of instances from clients participating in round \(t\).

   

\section{Evaluation}
\label{sec:evaluation}
\begin{figure}[t]
    \includegraphics[width=\linewidth]{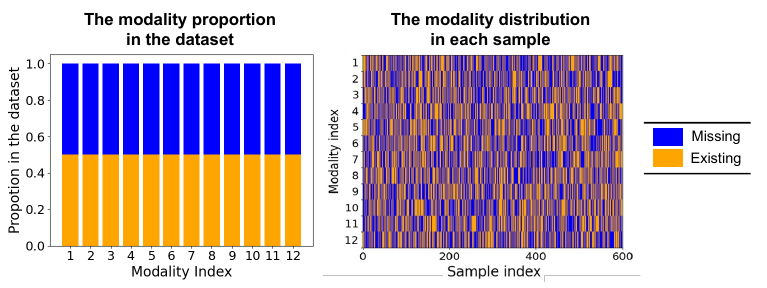}
     \caption{Example of an incomplete dataset \(\hat{D}\) with a missing pattern \((p_m, p_s) = (1.0, 0.5)\) indicates that 100\% of the modalities are randomly missing in 50\% of the samples.}
    \label{fig:client_modality_dist}
    \vspace{-10pt}
\end{figure}

\begin{figure*}[t]
\centering
\begin{subfigure}{\linewidth}
    \includegraphics[width=\textwidth]{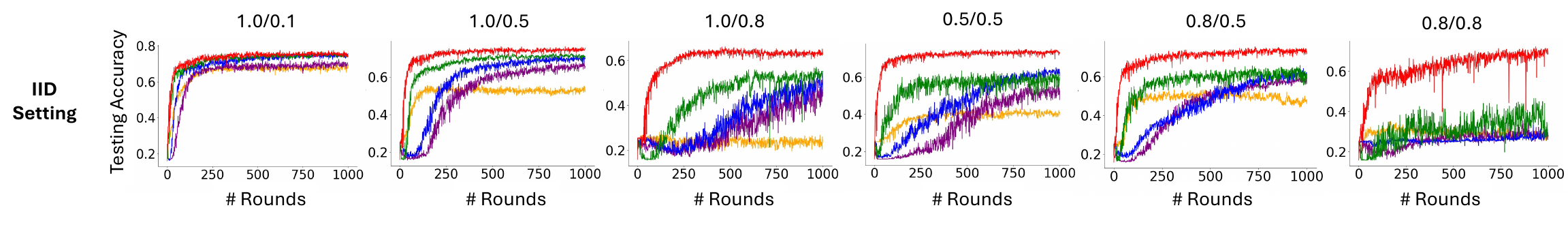}
\end{subfigure}
\hfill
\begin{subfigure}{\linewidth}
    \includegraphics[width=\textwidth]{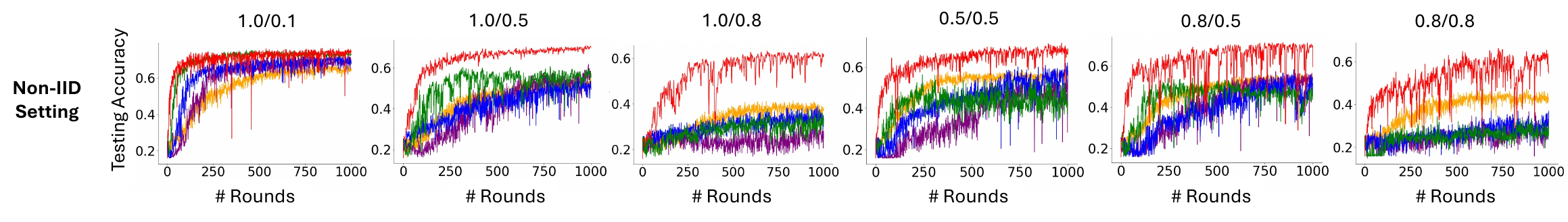}
\end{subfigure}
\vfill
\begin{subfigure}{0.55\linewidth}
    \includegraphics[width=\textwidth]{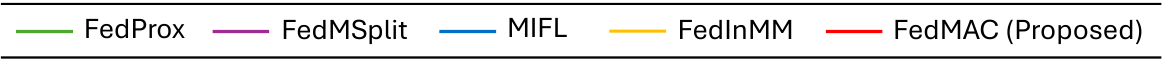}
\end{subfigure}
\caption{Performance of different methods over communication rounds under various missing statistics in the IID and Non-IID settings, where the missing statistics of the clients and server are \textit{similar}.}
\label{fig:test_plot}
\end{figure*}

\begin{table*}[t]
\centering
\caption{Comparison of FedMAC with other methods under various missing statistics in the IID and Non-IID settings, where the missing statistics of the clients and server are \textit{similar}.
The best and second-best results are highlighted in \textbf{bold} and \underline{underlined}, respectively.
}
\label{tab:iid_noniid_same}
{
\begin{tabular}{@{}l|cccccc||cccccc@{}}
\toprule
\multirowcell{2}[-2pt][l]{Model}  & \multicolumn{6}{c}{Missing statistics \((p_m/p_s)\) in the IID setting} & \multicolumn{6}{c}{Missing statistics \((p_m/p_s)\) in the Non-IID setting} \\  \cmidrule(lr){2-13}
 & {1.0/0.1} & {1.0/0.5} & {1.0/0.8} & {0.5/0.5} & {0.8/0.5} & {0.8/0.8} & {1.0/0.1} & {1.0/0.5} & {1.0/0.8} & {0.5/0.5} & {0.8/0.5} & {0.8/0.8} \\ \midrule
FedProx  & \underline{76.67} & \underline{72.51} & \underline{56.99} & 64.31 & 63.56 & \underline{46.53}  & \underline{75.16} & 60.53 & 37.96 & 56.24 & 35.06 & 34.43 \\
FedMSplit & 72.51 & 68.10 & 51.32 & 56.62      & 60.40      & 35.56  & 71.43 & 58.13 & 36.47 & 57.88      & 57.38      & 36.70    \\
MIFL  & 76.29 & 72.13 & 55.23 & \underline{65.19} & \underline{64.31} & 37.45 & 73.01 & \underline{64.43} & 38.97 & \underline{62.04} & \underline{58.51} & 36.95 \\
FedInMM   & 70.37 & 56.66 & 28.12 & 43.25  & 53.59 & 35.18 & 67.47 & 55.36 & \underline{41.36} & 59.64 & 56.37 & \underline{47.16}  \\ \midrule
\textbf{FedMAC}  & \textbf{77.42} & \textbf{76.16} & \textbf{66.96} & \textbf{75.03} & \textbf{75.28} & \textbf{72.27} & \textbf{76.29} & \textbf{74.40} & \textbf{64.82} & \textbf{72.38} & \textbf{73.39} & \textbf{67.59} \\ \bottomrule
\end{tabular}
}
\end{table*}

\subsection{Experimental Settings}
\noindent \textbf{Dataset and Preparation.}
Following Phung et al.~\cite{phung2024mifl}, we use a subset of the PTB-XL dataset~\cite{dataset}. This subset consists of 3,963 clinical samples across five distinct classes. 
Each sample has 12 modalities (the electrocardiogram  recordings) and is assigned to a single class. 
The dataset is subsequently divided into clients' training datasets and a server dataset with the 80:20 ratio.
The clients' dataset is then divided among $K$ clients according to the IID or Non-IID settings.

To simulate partial-modality missing data, the missing pattern \((p_m^k, p_s^k)\) for a client \(k\) with dataset \(D_k\) can be represented by a missing matrix:
\begin{equation}    
    \psi (D_k) = \begin{bmatrix}
                b_1^1 & \dots & b_1^{M_k}\\
                \vdots & \ddots & \vdots \\
                b_{N_k}^1 & \dots & b_{N_k}^{M_k}
                \end{bmatrix}, 
\end{equation}
where \( b_i^m \in \{0, 1\} \) indicates whether modality \(m\) is missing or available for sample \(i\). Here, \(M_k\) is the cardinality of \(\mathcal{M}(\mathbb{D}_k)\), and \(N_k\) is the number of samples. 
The incomplete dataset \(\hat{D}_k\) can be obtained by multiplying \(D_k\) by the missing matrix:
\begin{equation}
    \hat{x}_i = [x_i^1, \dots, x_i^{M_k}] \odot [b_i^1, \dots, b_i^{M_k}],
\end{equation}
where \(\odot\) represents element-wise multiplication.
Examples of incomplete dataset are shown in Fig.~\ref{fig:client_modality_dist}.

For simplicity, we refer to the missing pattern \((p_m^k, p_s^k)\) as \emph{missing statistics}, which reflects the statistical information about the partial-modality missing dataset. Additionally, we define the term \emph{missing degree} as \(p_m^k \times p_s^k\) to describe the overall amount of missing modalities in the entire dataset.

\noindent \textbf{Baseline Methods.} 
{
We compare the proposed FedMAC with several baseline methods, including FedProx~\cite{li2020federated}, FedMSplit~\cite{chen2022fedmsplit}, MIFL~\cite{phung2024mifl}, and FedInMM~\cite{yu2024fedinmm}. 
FedProx is a naive approach that does not explicitly address missing modalities, treating all instances equally regardless of their available modalities. 
FedMSplit and MIFL are methods designed to handle scenarios with complete-modality missing.  
On the other hand, FedInMM is the recently developed method for handling partial-modality missing scenarios.
}

\noindent \textbf{Evaluation Metric.}
{
We measure the effectiveness of all methods by calculating the accuracy of the server's dataset. 
This metric clearly indicates the model's overall performance after aggregating the learned parameters from all clients.
}

\noindent \textbf{Implementation Details.} 
We implement FedMAC as detailed in Sec.~\ref{sec:methodology}, utilizing an Inception Network as the Modality-Specific Extractor, as following~\cite{phung2024mifl}. 
For the classification task, we use Cross Entropy Loss for \(\mathcal{L}_{task}\). 
The embedding dimension is set to \(d_H = 128\). The number of clients is \(K = 32\), as following~\cite{yu2024fedinmm}, with local training epochs \(E = 3\), a batch size \(B = 32\), and a learning rate of \(\eta = 0.01\) for the IID setting and \(\eta = 0.008\) for the Non-IID setting. 
The optimizer used is Stochastic Gradient Descent (SGD)~\cite{ruder2016overview}. 
Communication with the server occurs over \(T = 1000\) rounds. 
The parameter \(\lambda\) in Eq.~\ref{eq:final_loss} is set to $0.1$ for \(p_m \times p_k \leq 0.5\) and $0.2$ otherwise. 
The temperature hyperparameter in Eqs.~\ref{eq:cross-modal-agg},~\ref{eq:contrasitve_shared_loss}, and~\ref{eq:contrastive_sim_loss} are set to $1$. 
For the other methods, we employ the configurations as specified in their original papers.

\subsection{Experimental Results}
{
To evaluate the effectiveness of FedMAC, we consider two scenarios: one where the missing statistics of the clients and server are similar, and another where the missing statistics between the clients and server differ. 
The former represents a basic setting with consistent missing statistics, while the latter reflects a more realistic scenario where the missing statistics of the clients and servers are different.
}

\subsubsection{Similar missing statistics between the client and server}
{
In this experiment, we set up different missing patterns under IID and Non-IID settings. 
IID refers to the scenario where the samples across all clients are identical, while Non-IID refers to the scenario where the samples differ between clients. 
We generate the Non-IID setting using a Dirichlet distribution, formulated as \( \pi \sim \text{Dirichlet}(\alpha) \), where \( \alpha \) is the concentration parameter. We set \( \alpha = 0.9 \) to indicate slight skewness, as the focus is on addressing modality heterogeneity among clients rather than skewed distributions.
}



\label{subsec:iid_eval}

\noindent \textbf{IID Setting Results.}
{
Fig.~\ref{fig:test_plot} and Table~\ref{tab:iid_noniid_same} present the results of the proposed FedMAC compared to other methods. 
The proposed method consistently outperforms the others across all experimented scenarios with various missing statistics. 
When the missing degree is low (e.g., missing statistic \(p_m/p_s = 1.0/0.1\), resulting in a missing degree of \(p_m \times p_s = 0.1\)), the performance differences between FedMAC and the other methods are minimal, with all achieving similar accuracy levels. 
However, as the missing degree increases (e.g., \(p_m \times p_s = 0.5\) and \(p_m \times p_s = 0.8\)), FedMAC demonstrates a significant advantage, maintaining higher accuracy than its counterparts. 
This trend becomes even more pronounced in more complex scenarios where the missing statistics of the clients are intricate, such as \(0.5/0.5\), \(0.8/0.5\), and \(0.8/0.8\). 
While other methods experience more significant performance degradation, FedMAC remains robust and consistently achieves the highest accuracy in every case.
These results underscore the ability of FedMAC to effectively manage modality heterogeneity, particularly as the missing degree increases in the IID setting.
}

\begin{table}[t]
\centering
\caption{Comparison of FedMAC with other methods under various missing statistics in the IID setting, where the missing statistics of the clients and server are \textit{different}.
The best and second-best results are highlighted in \textbf{bold} and \underline{underlined}, respectively.}
\label{tab:iid_different}
\setlength\tabcolsep{4pt} 
\resizebox{\linewidth}{!}{%
\begin{tabular}{@{}c|c|l|cccccc@{}}
\toprule
& & \multirowcell{2}[-2pt][l]{Model} & \multicolumn{6}{c}{Server's missing statistics $(p_m/p_s)$} \\ \cmidrule(lr){4-9}
& & & {1.0/0.1}   & {1.0/0.5}   & {1.0/0.8}   & {0.5/0.95}   & {0.8/0.95}   & {0.1/0.95}   \\ \midrule
\multirow{17}{*}{\rotatebox{90}{Client's missing statistics $(p_m/p_s)$}} & \multirow{5}{*}{\rotatebox{90}{1.0/0.1}} & FedProx        & 76.54          & 68.72          & 51.83    & 70.49          & 51.83          & {\ul 76.54}    \\
& & FedMSplit & 72.76 & 66.33 & 52.33  & 69.99 & 59.27       & 72.89 \\
& & MIFL & {\ul 76.67}    & {\ul 71.00}    & \textbf{53.85}  & {\ul 72.01} & {\ul 61.79}    & 76.04          \\
& & FedInMM   & 68.73 & 57.63 & 43.76   & 64.94 & 55.11       & 71.75 \\ \cmidrule(lr){3-9} 
&  & \textbf{FedMAC}  & \textbf{77.93} & \textbf{72.51} & {\ul 52.46}    & \textbf{75.53} & \textbf{65.83} & \textbf{77.30} \\ \cmidrule(lr){2-9} 
& \multirow{5}{*}{\rotatebox{90}{1.0/0.5}} & FedProx & \textbf{75.53} & {\ul 72.13}    & 60.40    & {\ul 72.76}    & 65.83          & {\ul 75.03}  \\
& & FedMSplit & 72.13 & 67.59 & 59.39      & 69.10 & 64.44       & 71.88 \\
& & MIFL& 74.65 & 71.63 & {\ul 60.78} & 71.25 & {\ul 66.08} & 72.76 \\
& & FedInMM   & 63.05 & 57.50 & 45.77  & 57.88 & 54.48       & 64.56 \\ \cmidrule(lr){3-9} 
& & \textbf{FedMAC}         & {\ul 75.16}    & \textbf{76.16} & \textbf{66.71} & \textbf{74.15} & \textbf{73.27} & \textbf{77.30} \\ \cmidrule(lr){2-9} 
& \multirow{5}{*}{\rotatebox{90}{1.0/0.8}} & FedProx        & {\ul 69.98}    & {\ul 66.46}    & {\ul 56.12}  & {\ul 69.23}    & {\ul 64.19}    & {\ul 69.36}    \\
& & FedMSplit & 65.32 & 60.91 & 49.31    & 65.32 & 57.63       & 66.83 \\
& & MIFL      & 68.35 & 64.69 & 54.22   & 66.71 & 57.00       & 67.59 \\
& & FedInMM   & 31.40 & 25.98 & 25.98     & 38.46 & 26.10       & 31.02 \\ \cmidrule(lr){3-9} 
& & \textbf{FedMAC} & \textbf{75.28} & \textbf{74.02} & \textbf{66.96} & \textbf{74.91} & \textbf{74.27} & \textbf{75.28} \\ \bottomrule
\end{tabular}%
}
\end{table}
\noindent \textbf{Non-IID Setting Results.}
{
As shown in Fig.~\ref{fig:test_plot} and Table~\ref{tab:iid_noniid_same}, FedMAC outperforms all other methods. 
While other methods experience significant drops in accuracy as the missing degree increases, the accuracy of FedMAC remains consistently higher. 
In challenging settings with high missing degrees, such as $1.0/0.8$, $0.8/0.5$, and $0.8/0.8$, FedMAC maintains a clear advantage, demonstrating its robustness and adaptability to non-IID data distributions. 
However, in the non-IID setting, each client's data represents a different distribution, leading to inconsistencies in the accuracy of FedMAC when aggregating model updates at the server during each communication round. 
The reason for this is that we use the FedAvg algorithm~\cite{mcmahan2017fedavg} for server aggregation in these experiments, which struggles in the non-IID setting because it aggregates based on the number of client samples. 
In future work, we are considering more robust aggregation strategies to better handle the variability in client data distributions and improve overall model performance in non-IID scenarios.
}

\subsubsection{Different missing statistics among client and server}
The results presented in Table~\ref{tab:iid_different} highlight the effectiveness of FedMAC in scenarios where the missing statistics between the clients and the server differ, which closely mirrors real-world conditions. 
FedMAC consistently outperforms other methods across various settings, demonstrating its robustness, particularly when the server's missing statistics become more severe, such as in the $1.0/0.5$, $1.0/0.8$, and $0.8/0.95$ cases. 
While FedProx and MIFL perform relatively well when the clients and server have lower missing degrees, these methods struggle as the server's missing degree increases.
In contrast, FedMAC maintains strong performance, consistently achieving the best results across nearly all experiments.
Notably, in scenarios with severe missing degrees on the server side (i.e., $1.0/0.8$ and $0.8/0.95$), the accuracy of FedMAC remains high, while FedInMM performs poorly under these challenging conditions. 
These results underscore the ability of FedMAC to effectively handle diverse and complex missing statistics, making it a reliable method for FL environments where data availability differs significantly between clients and the server.

\begin{table}[t]
\centering
\caption{Ablation study on different components of FedMAC under various missing statistics in the IID setting, where the missing statistics of the clients and server are \textit{similar}.
The best and second-best results are highlighted in \textbf{bold} and \underline{underlined}, respectively.}
\label{tab:abl_study}
{%
\begin{tabular}{@{}l|cccccc@{}}
\toprule
\multirowcell{2}[-2pt][l]{Model} & \multicolumn{6}{c}{Missing statistics \((p_m/p_s)\)} \\ \cmidrule(lr){2-7}
 & {1.0/0.1} & {1.0/0.5} & {1.0/0.8} & {0.5/0.5} & {0.8/0.5} & {0.8/0.8} \\ \midrule
FedC & \textbf{77.80} & \underline{76.08} & \textbf{67.09} & \underline{73.26} & \underline{74.40} & \underline{68.22} \\
FedMA & 73.52 & 72.76 & 66.82 & 61.69 & 66.96 & 61.64 \\ \midrule
\textbf{FedMAC} & \underline{77.42} & \textbf{76.16} & \underline{66.96} & \textbf{75.03} & \textbf{75.28} & \textbf{72.26} \\ \bottomrule
\end{tabular}%
}
\end{table}
\subsection{Ablation Study}
{
In this section, we evaluate two key components of the proposed FedMAC: Cross-Modal Aggregation and Contrastive Regularization, under various missing statistics in the IID setting, where the missing statistics of the clients and server are similar.
We implement the following two variants:
\begin{itemize}
    \item \textbf{FedC:} FedMAC without Cross-\textbf{M}odal \textbf{A}ggregator, which is designed to capture inter-modal information and reduce potential biases toward specific modalities.
    \item \textbf{FedMA:} FedMAC without \textbf{C}ontrastive Regularization. We exclude $\mathcal{L}_{sim}$ and $\mathcal{L}_{shared}$ from the training loss function in Eq.~\ref{eq:final_loss} to evaluate the impact of the multi-modal contrastive-based constraint on similarity learning.
\end{itemize}
}
Table~\ref{tab:abl_study} presents the results of FedMAC and its ablation variants, FedC and FedMA. 
The findings highlight the importance of Cross-Modal Aggregation and Contrastive Regularization in enhancing the performance of FedMAC under various missing modality scenarios. 
FedMAC consistently outperforms FedC and FedMA across most scenarios. 
FedC, which lacks Cross-Modal Aggregation, performs adequately when the missing modality pattern is uniform across clients (\(p_m = 1\)) but struggles when the pattern varies (\(p_m < 1\)), indicating that aggregation is vital for capturing inter-modal relationships. 
FedMA, which omits Contrastive Regularization, shows notable performance declines, particularly in scenarios with high client variation (\(p_m < 1\)), underscoring the role of contrastive mechanisms in maintaining modality-invariant representations. 
Overall, the ablation study demonstrates that the combination of Cross-Modal Aggregation and Contrastive Regularization in FedMAC is crucial for robust and consistent performance, ensuring the model's adaptability in diverse and challenging scenarios.

\section{Conclusion}
\label{sec:conclusion}
This study proposed a novel multi-modal FL framework named FedMAC, designed to address modality heterogeneity among clients with instance-level incompleteness effectively. 
The proposed method adopts a Cross-Modal Aggregator to strengthen inter-modal relationships and mitigate modality bias, while the Contrastive-based Regularization ensures more accurate and consistent model updates during client training. 
Empirical results demonstrate the superior performance of the proposed method across various challenging scenarios, highlighting its ability to adapt to different patterns of missing data. 
This adaptability positions FedMAC as a robust and versatile solution for real-world multi-modal FL applications, ultimately contributing to more equitable and effective deployment of FL in diverse environments. 
Future work will focus on server aggregation design and extensive experiments on diverse multi-modal datasets. 

\section{Acknowledgement}
This research is funded by Hanoi University of Science and Technology (HUST) under grant number T2023-PC-028.
This work was funded by Vingroup Joint Stock Company (Vingroup JSC),Vingroup, and supported by Vingroup Innovation Foundation (VINIF) under project code VINIF.2021.DA00128.

\bibliographystyle{IEEEtran}
\bibliography{references}

\end{document}